\def\year{2022}\relax
\documentclass[letterpaper]{article} 
\usepackage{aaai22}  
\usepackage{times}  
\usepackage{helvet}  
\usepackage{courier}  
\usepackage[hyphens]{url}  
\usepackage{graphicx} 
\usepackage{subfigure}
\usepackage{amsmath}
\usepackage{amsfonts,amssymb}
\usepackage{multirow}
\usepackage{natbib}  
\usepackage{caption} 
\DeclareCaptionStyle{ruled}{labelfont=normalfont,labelsep=colon,strut=off} 
\usepackage{algorithm}
\usepackage{algorithmic}
\usepackage{float}
\usepackage{amsmath,amssymb}
\usepackage[linewidth=1pt]{mdframed}
\usepackage[colorlinks=true,linkcolor=red,anchorcolor=black,colorlinks=false,citecolor=black,CJKbookmarks=false]{hyperref}
\usepackage{appendix}
\usepackage{array}

\usepackage{newfloat}
\usepackage{listings}
\floatstyle{ruled}
\newfloat{listing}{tb}{lst}{}
\floatname{listing}{Listing}
\title{OctAttention: Octree-Based Large-Scale Contexts Model for\\Point Cloud Compression}
\author{
    Chunyang Fu\textsuperscript{\rm 1}\textsuperscript{,}\textsuperscript{\rm 2},
    Ge Li\textsuperscript{\rm 1},
    Rui Song\textsuperscript{\rm 1},
    Wei Gao\textsuperscript{\rm 1}\textsuperscript{,}\textsuperscript{\rm 2}\thanks{Corresponding author},
    Shan Liu\textsuperscript{\rm 3}
}
\affiliations{
    \textsuperscript{\rm 1}School of Electronic and Computer Engineering, Peking University Shenzhen Graduate School\\
    \textsuperscript{\rm 2}Peng Cheng Laboratory\\
    \textsuperscript{\rm 3}Tencent America\\
    fuchy@stu.pku.edu.cn, geli@ece.pku.edu.cn, rsong@stu.pku.edu.cn, gaowei262@pku.edu.cn,  shanl@tencent.com

}
\begin{document}
\maketitle

\begin{abstract}
In point cloud compression, sufficient contexts are significant for modeling the point cloud distribution. However, the contexts gathered by the previous voxel-based methods decrease when handling sparse point clouds. To address this problem, we propose a multiple-contexts deep learning framework called \emph{OctAttention} employing the octree structure, a memory-efficient representation for point clouds. Our approach encodes octree symbol sequences in a lossless way by gathering the information of sibling and ancestor nodes. Expressly, we first represent point clouds with octree to reduce spatial redundancy, which is robust for point clouds with different resolutions. We then design a conditional entropy model with a large receptive field that models the sibling and ancestor contexts to exploit the strong dependency among the neighboring nodes and employ an attention mechanism to emphasize the correlated nodes in the context. Furthermore, we introduce a \emph{mask} operation during training and testing to make a trade-off between encoding time and performance. Compared to the previous state-of-the-art works, our approach obtains a 10\%-35\% BD-Rate gain on the LiDAR benchmark (\emph{e.g.}  SemanticKITTI) and object point cloud dataset (\emph{e.g.} MPEG 8i, MVUB), and saves 95\% coding time compared to the voxel-based baseline. The code is available at \url{https://github.com/zb12138/OctAttention}.

\end{abstract}

\section{Introduction}
The point cloud is an essential data structure for 3D representation. It has been used in many fields such as virtual reality, smart city, robotics, and autonomous driving \cite{23}. Since massive point clouds are generated, efficient compression techniques are necessary for point cloud storage and transmission. However, point clouds are unordered and have various distributions; it is relatively difficult to compress point clouds compared with 2D images. Fortunately, several schemes for point cloud geometry and attribute compression such as
voxel-based, image-based and tree-based algorithms have been proposed and applied in
research works \cite{chou2019volumetric,shao2017attribute,shao2018hybrid} and
standard specification \cite{gpcc}.

\begin{figure*}[!ht]
\centering
 \subfigure{\includegraphics[scale=0.26]{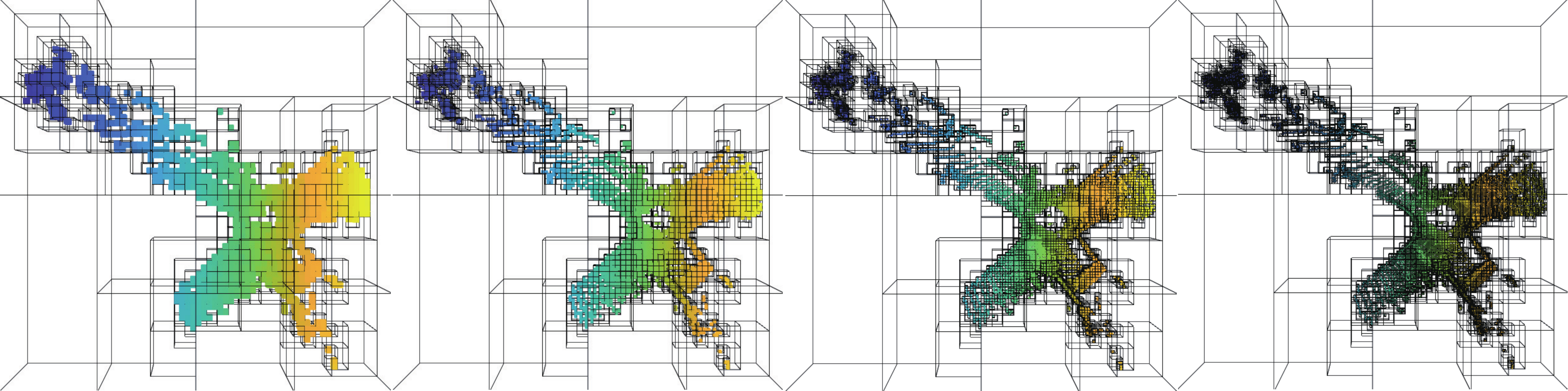}}
 \caption{LiDAR point cloud quantized by octree with the max depth of 6, 7, 8, 9. Points in the octree leaf nodes decrease with increased depth. The receptive field of voxel-based approach is limited.
}
\label{Octree}
\end{figure*}

The MPEG point cloud geometry compression standard (G-PCC) \cite{23} adopted a hand-crafted context-adaptive arithmetic encoder for bit allocation, which can be seen as a prediction for the currently encoding node based on the coded information. Recently, entropy encoders based on deep learning have been shown to outperform hand-crafted ones on rate-distortion performance. Among them, some methods partition point clouds into voxels, then adopt 3D convolution to learn and predict the occupancy of each voxel \cite{33,voxeldnn,27,que2021VoxelContext}. Voxel-based models are capable of exploiting the local geometric patterns (\emph{e.g.}, planes, surfaces). However, they are not robust to point cloud resolution. These methods have to divide blocks into different scales for point clouds with varying point densities to find the optimal voxel size.
Meanwhile, receptive fields are limited by the computational cost, \emph{i.e.}, they only extract features from voxels within a narrow range. Other works encode the point cloud into an octree, then encode the occupancy of octree nodes based on their ancestor nodes \cite{huang2020octsqueeze,biswas2021muscle}. The octree-based model is robust to resolution, and it also utilizes a broader range of contexts than voxel-based ones. However, prior methods ignore that sibling nodes (\emph{i.e.}, nodes in the same octree level) provide low-level local geometry features, which are significant to exploit the geometry redundancy. In general, prior voxel-based and octree-based methods do not fully use much spatial context information.

In this work, we propose a point cloud compression method called \emph{OctAttention} which generates and utilizes a large-scale context. Voxel is inefficient for representing sparse point clouds; thus, we encode the point cloud into an octree. Subsequently, we improve the predictability of the occupancy of each octree node based on large-scale context, which contains features from ancestor nodes of the current node, sibling nodes, and ancestors of sibling nodes. However, it should be noted that the side effect of expanding context is to introduce redundant and irrelevant information. For example, different sibling nodes may have the same ancestors, but they are repeated multiple times in the context; sibling nodes far from the current node may be worthless for prediction. To tackle this problem, we employ tree-structured attention to weight and explicitly express the contributions of different nodes in the prediction. The siblings in the context disable the parallelization strategy in prior works \cite{huang2020octsqueeze,que2021VoxelContext}, hence we propose a \emph{mask} operation to encode multiple nodes in parallel.

We compare the proposed model with state-of-the-art methods on the 3D LiDAR dataset SemanticKITTI \cite{semantickiiti}, object point cloud dataset MVUB \cite{MVUB} and MPEG 8i \cite{MPEG8i}. The experiments show that our method outperforms these state-of-the-art methods, which are only designed for a specific category of point clouds.

The contributions of our work can be summarized as:
\begin{itemize}
\item We propose a tree-structured attention mechanism to model the dependency of nodes in a large-scale context, which is achieved by extending the receptive field of context and exploiting features from sibling nodes and their ancestors.
\item We employ a \emph{mask} operation to encode octree in parallel to alleviate the drawbacks of introducing siblings in the large-scale context.
\item Our generic model of point cloud geometry compression for both LiDAR and object point clouds achieves state-of-the-art performance on several large-scale datasets.
\end{itemize}

\section{Related Work}
\subsection{Voxel-Based Point Cloud Compression}
Voxel-based methods quantize the point cloud and classify the voxel occupancy by neural networks. Voxel-based methods outperform G-PCC \cite{gpcc} on lossless geometric compression \cite{voxeldnn,27}, lossy geometric compression \cite{27,28,33} and progressive compression \cite{29}. Compared to an octree, geometric patterns can be naturally preserved in the voxel representation. Yet, the side effect is voxel-based networks are sensitive to the density variation and may fail for the sparse point clouds. All of the above methods are applied to dense point clouds (\emph{e.g.} MPEG 8i) and may suffer tremendous computing and memory costs on sparse LiDAR point clouds. The proposed method directly processes the octree occupancy code to overcome the density variation problem.
\subsection{Tree-Based Point Cloud Compression}
Tree structures effectively reduce geometric redundancy by merging the common coordinates of point clouds. Numerous algorithms \cite{3,5,23,9,11,12} compressed point cloud based on tree structures such as octree \cite{16}, quadtree \cite{6}, KD tree\cite{2}, prediction tree \cite{10}, \emph{etc.} Recently, many works have focused on designing octree context for arithmetic coding to compress bitstream.
\cite{song2021layer} aggregated voxels in reduced space by removing free regions and acquires a compact context model for LiDAR compression. \cite{8} reordered the node sequences to improve the intra-frames lossless geometric compression.

All of the above methods model the context by hand-crafted features. \cite{01,02,03} mainly introduce new convolution methods under the octree framework. OctSqueeze \cite{huang2020octsqueeze} is the first octree-based deep learning entropy model by modeling the dependency among node and its multiple ancestor nodes. MuSCLE \cite{biswas2021muscle} reduced the temporal redundancy by exploiting spatio-temporal relationships across LiDAR sweeps. Both methods avoid high computational complexity, yet the strong dependency among sibling nodes is ignored. VoxelContext-Net \cite{que2021VoxelContext} partly solved this problem by employing voxel and octree hybrid structure to learn the context in the previous octree depth. However, features from higher resolution (\emph{i.e.}, from sibling nodes) are still ignored. Besides, as shown in Fig. \ref{Octree}, given a fixed-size voxel-based context, its receptive field shrinks with the increased octree depth. Due to the computational overhead, the receptive field of the voxel-based approach is limited, which restricts the ability to model the context. Our proposed method can acquire more than $500\times500\times500$ voxels receptive field ($9\times9\times9$ in VoxelContext-Net). Meanwhile, we introduce sibling nodes and their ancestors in the context. The extended context contains more potentially helpful information to model the distribution of octree nodes.

\begin{figure*}
\centering
 \subfigure{\includegraphics[scale=0.331]{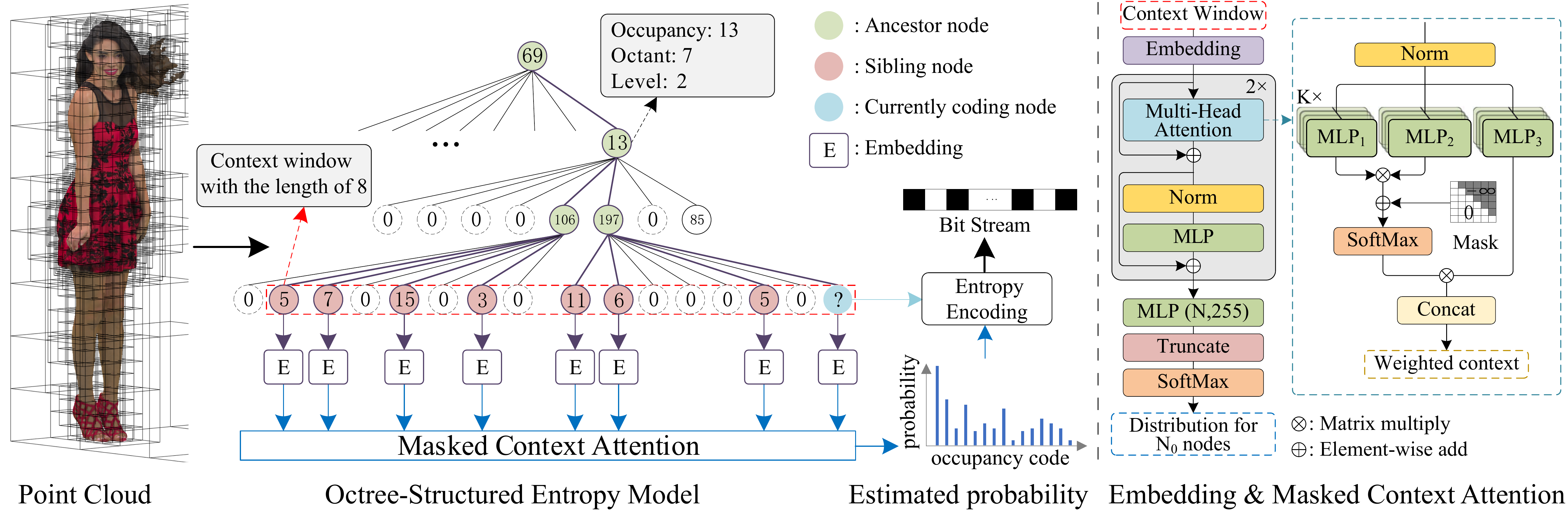}}
 \caption{The overview architecture of \emph{OctAttention} (Left). The number in the node indicates the corresponding occupancy code in decimal. The point cloud is first encoded into an octree, in which an octree node is characterized by its occupancy code, level, and octant. These 3 features are embedded separately. For example, we construct a context window (red) with length $N=8$. The involvement of 3 layers of ancestors in the context window is marked green (\emph{i.e.} the context window height $K=4$). While encoding a node (blue), the context in the window is then fed into a masked context attention module (Right) and eventually utilized to model the occupancy code distribution by multi-layer perceptron (MLP). Finally, we use the estimated distribution to encode the serialized occupancy code into the final compressed bit stream by the arithmetic encoder.
}
\label{frameWork}
\end{figure*}

\section{Methodology}
We propose an extended context and a tree-structured attention mechanism, which is shown in Fig. \ref{frameWork}. Intuitively, nodes with similar ancestors and siblings tend to follow a similar distribution. Therefore, the proposed context exploits features from siblings and their ancestors, which are beneficial for inference. To achieve accurate and flexible prediction with a large-scale context, we employ a tree-structured attention mechanism to determine the importance of each node in the context. Finally, we infer the occupancy of each octree node based on the attention context. We further propose a \emph{mask} operation to counteract the increased coding time caused by the involvement of the sibling context.
\subsection{Octree Structure}
Octree is an efficient approach to organizing a point cloud (see Fig. \ref{Octree}). Given a point cloud ${P}$, we translate the point cloud by an $\text{offset}= (\operatorname{min}(P_\mathrm{x}),\operatorname{min}(P_\mathrm{y}),\operatorname{min}(P_\mathrm{z}))$, and quantize it by quantization step $\mathrm{qs}$:
\begin{align}
&P_{\mathrm{Q}}=\operatorname{round}\left(\frac{P-\text { offset }}{\mathrm{qs}}\right) \label{PQ}
\end{align}
\begin{align}
&\mathrm{qs} \geq \frac{\max (P)-\min (P)}{2^{\mathrm{L}}-1}\label{qs}
\end{align}
An octree divides the cube space along the maximum side length of the bounding box of ${P}_{\mathrm{Q}}$ into 8 equal octants recursively. The occupancy status of 8 children cubes constitutes an 8-bit binary occupancy code. Only the nonempty child cubes will be flagged as 1 and be further subdivided. Other unoccupied cubes will be flagged as 0. The division will be terminated at the given depth $\mathrm{L}$. At the leaf nodes, an 8-bit occupancy code represents 8 cubes whose side length equals the quantization step $\mathrm{qs}$. Points in the $P$ are aligned and merged with the nearest corresponding cube. The point cloud is reconstructed by inverse quantization: $\hat{P}=P_Q*\mathrm{qs}+\text {offset}$. Given the reconstructed point cloud $\hat{P}$, the reconstruction error will be controlled in:
\begin{align}
e=\max_{i}||\hat{P}_{{i}}-{P}_{i}||_{\infty} \leq \frac{\mathrm{qs}}{2} \label{DQerror}
\end{align}
The geometric compression loss of our method only comes from quantization error in Eq. (\ref{DQerror}). Therefore, we are allowed to increase the octree depth $\mathrm{L}$ to achieve the arbitrary required accuracy.
\subsection{Context Model}
According to Shannon's source coding theorem \cite{shannon2001mathematical}, given an octree in breadth-first search fashion $\mathbf{x} =\{x_{1},\ldots,x_{i},\ldots\}$, its bitrate is lower bounded by its information entropy. Since the ground truth distribution $P(\mathbf{x})$ is high-dimensional and unknown, we estimate a probability $Q(\mathbf{x})$ by a network with prior information to guide the arithmetic coder. The closer the estimated probability is to ground truth, the closer the actual bitrate $\mathbb{E}_{\mathbf{x} \sim P}\left[-\log _{2} Q(\mathbf{x})\right]$ is to its lower bound $\mathbb{E}_{\mathbf{x} \sim P}\left[-\log _{2} P(\mathbf{x})\right]$. Therefore, we aim to estimate accurate occupancy distribution to reduce the bitrate. We believe that exploiting more features in the context benefits this goal. Large-scale context is significant in extending the receptive field of context and searching for dependency in a wide range. Sibling nodes also serve as a complement of low-level local geometry features. For example, leaf sibling nodes are likely to distribute on the neighboring surface in the point cloud and have similar geometry patterns, which the sibling context can capture.

We propose an expanded large-scale context to achieve more accurate probability estimation. We first traverse the octree in a breadth-first order. Then for each currently encoding node $n_i$ in the sequence, we construct a context window  
$\left\{{n}_{i-N+1},\ldots, {n}_{k},\ldots,{n}_{i-1},{n}_{i} \right\}$ with the length of $N$ (see Fig. \ref{frameWork}  Left). The currently encoding node is at the end of the window, and the local context window slides forward with the currently encoding node moving forward. In this manner, we select $N-1$ sibling nodes to exploit the strong dependency among the nodes at the same depth. Considering the dependency between the nodes and their ancestor nodes, we further introduce $K-1$ ancestors of the $N$ nodes in the context window, respectively. In summary, we integrate $N*K$ related nodes and greatly expand the contexts. Specifically, we factorize distribution $Q(\mathbf{x})$ into a product of conditional probabilities of each occupancy symbol $x_{i}$ as:
\begin{align}
Q(\mathbf{x})=\prod_{i}q_i\left(x_{i} \mid \boldsymbol{f}_{i-N+1},\ldots,\boldsymbol{f}_{k},\ldots,\boldsymbol{f}_{i};\mathbf{w}\right) \label{Q(x)}
\end{align}
where $x_{i}\in\{1,2, \ldots, 255\}$ is the occupancy symbol of currently encoding node $n_i$. High-dimensional vector $\boldsymbol{f}_k=[\boldsymbol{h}_k^{(0)},\boldsymbol{h}^{(1)}_{k},\boldsymbol{h}^{(2)}_{k},...,\boldsymbol{h}^{(K-1)}_{k}]$ denotes the concatenation of embedding features of nodes $n_k$ and features of its $K-1$ ancestors, which are defined as the embedding of their occupancy, depth and octant. $\mathbf{w}$ denotes the context model parameters.

Voxel-based methods naturally exploit the low-level geometry features preserved by a voxel representation. However, in an octree, these low-level features are hidden in the octree nodes. We employ an embedding for each node in the context window to reveal these features. Node information is embedded to a fixed-length vector respectively and then concatenated as $\boldsymbol{h}_{k}=\left[S_kW_{1}, L_kW_{2},O_kW_3\right]$, where $S_k,L_k,O_k$ are one-hot coded occupancy, level index and octant index, and $W$ is their respective embedding matrix. Embedding also serves as normalization for the three different scale variables. It should be noted that the depth and octant of $n_i$  are already available while decoding $n_i$, yet its occupancy code $x_i$ is unknown, so we pad it with $x_{i-1}$. 

In this manner, we utilize the features from sibling nodes and their ancestors, which are significant to prediction. Previous octree-based works \cite{huang2020octsqueeze,biswas2021muscle} ignored it. In these works, node $n_i$ is assumed to be conditional independence under the condition $\boldsymbol{f}_i$. We strengthen this condition by expanding the context. Thus we make a more general assumption. Meanwhile, we avoid the inefficient sparse context and tremendous incremental computations caused by expanding the context in previous voxel-based works \cite{que2021VoxelContext}. The receptive field of the proposed context can exceed 1000 octree nodes, which is extremely difficult to be achieved in voxel-based methods.

\subsection{Tree-Structured Attention}
By expanding the context, more information is available for inference. However, a portion of nodes in the context is worthless. See Fig. \ref{frameWork} (Right). We adopt the self-attention mechanism \cite{18} to discover similarity and strong dependency among the nodes, filter out irrelevant nodes,  and emphasize useful ones. Since self-attention effectively tackles long-range dependence, it is also appropriate in a large-scale context. Meanwhile, attention achieves flexible inference by evaluating weights varying with the input. 
Intuitively, nodes having similar ancestors and siblings tend to follow similar distribution, hence it is reasonable to estimate the occupancy based on similarity among siblings to the estimated node $n_i$. We omit shortcuts and normalization for brief illustration. In the head $t$ ($t=0,1,...,K-1$), the attention score scalar in attention map between the ${m}^\mathrm{{th}}$ and ${n}^\mathrm{{th}}$ sibling node in the context window is defined as:
\begin{align}
a_{m,n}^{(t)} = \frac{\operatorname{exp}(\text{MLP}_\mathrm{1}(\boldsymbol{h}_{m}^{(t)}) \cdot
\text{MLP}^\mathrm{{T}}_\mathrm{{2}}(\boldsymbol{h}_{n}^{(t)}) )}{\sum_{k=i-N+1}^{m}\operatorname{exp}(\text{MLP}_\mathrm{1}(\boldsymbol{h}_{m}^{(t)}) \cdot \text{MLP}^\mathrm{{T}}_\mathrm{{2}}(\boldsymbol{h}_{k}^{(t)}))}
\label{att}
\end{align} 
where $n \leq m, m={i-N+1,...,i}$. The summation for node $n_m$ ends at $m$ since a \emph{mask} operation is applied to the attention map to achieve fast encoding, which is discussed in the next section. With attention mechanism, we can draw weighted context as:
\begin{align}
\boldsymbol{C}_i^{(t)}= \sum_{k=i-N+1}^{i} a_{i,k}^{(t)} \cdot \text{MLP}_\mathrm{3}(\boldsymbol{h}_{k}^{(t)})
\end{align} 
In summary, the contexts are fed to 2 layers of multi-head self-attention $\text{MultiHead}(\boldsymbol{f},i) =\text{MLP}([\boldsymbol{C}_i^{(0)},...,\boldsymbol{C}_i^{(K-1)}]) $ and multi-layer perception (MLP) successively, and finally outputs a 255-dimensional probability for $n_{i}$:
\begin{align}
q_i\left(\cdot\mid\boldsymbol{f};\mathbf{w}\right)=\text{softmax}(\text{MLP}(\text{MultiHead}^{(2)}(\boldsymbol{f},i)))
\label{qx}
\end{align} 
\begin{figure}[!b]
\centering
 \subfigure[]{\includegraphics[scale=0.41]{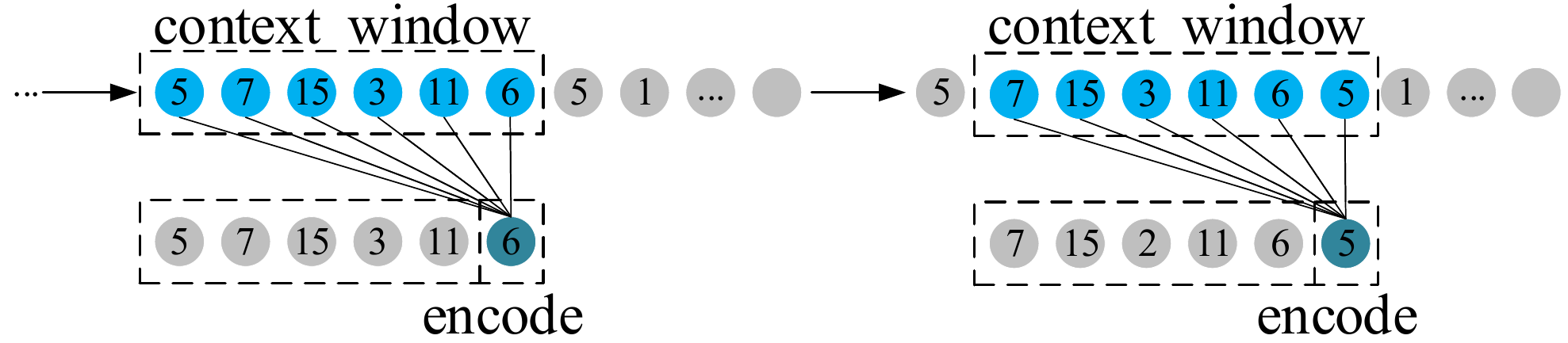}}
  \subfigure[]{\includegraphics[scale=0.41]{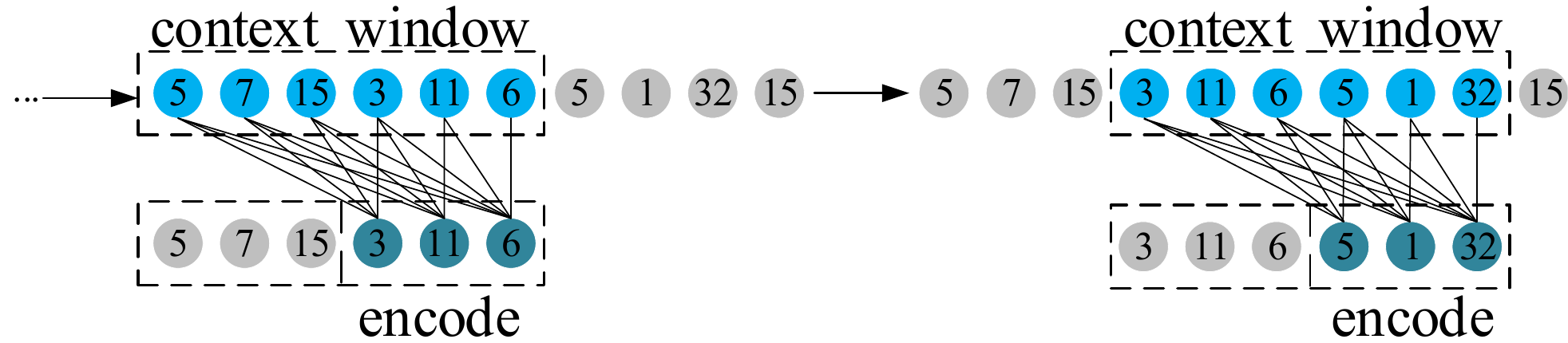}}
 \caption{Illustration of \emph{mask} operation. The vertical line represents the involvement of octant and level information. (a) Without \emph{mask}. $N_{0}=1$ and receptive field of encoded nodes is 6. (b) With \emph{mask}. $N_{0}=3$ and receptive fields of encoded nodes are 4,5,6. Meanwhile, it reduces encoding time by 3 times compared to (a).
}
\label{slindeWin}
\end{figure}

\subsection{Mask operation}
The estimated probability is adopted to guide the arithmetic coder, which codes the octree nodes sequentially in a lossless way. Previous methods \cite{huang2020octsqueeze,que2021VoxelContext} excluded siblings from contexts and only depend on ancestors. Hence they can naturally parallelize encoding and decoding within each level. Without \emph{mask} operation, as shown in Fig. \ref{slindeWin}(a), given a sliding window, we adopt all $N-1$ siblings in the context to predict the last node in the context window. It is difficult to achieve the same parallelization since we can only encode the last node in one propagation. To reduce encoding time, we introduce a \emph{mask} operation in Eq. (\ref{att}) which assigns a varied-length receptive field for each node. Each node is restricted to the access of the previous nodes in the context window at training and testing. As shown in Fig. \ref{slindeWin}(b), in this way, Eq. (\ref{qx}) can equally apply to the last $N_{0}$ nodes. Hence we are allowed to encode them simultaneously in one propagation. While encoding the $j^{\text{th}}$ node in the window, only $j$ nodes are available for inference. Compared to the way of maximum receptive field in Fig. \ref{slindeWin}(a), on average, the receptive field of each node shrinks from $N$ to $(2N-N_0+1)/2$. Nevertheless, the coding time reduces by $N_{0}$ times. The parameter $N_{0}$ balances the receptive field and coding time. Although the receptive field shrinks, we find there is negligible performance loss as the result of the \emph{mask} operation during the training. 

\subsection{Learning}
We optimize the cross-entropy between the predicted occupancy code and ground-truth, which is defined as:
\begin{equation}
\ell=-\sum_{i} \log q_i\left(x_{i} \mid \boldsymbol{f};\mathbf{w} \right)
\end{equation}
Here, $q_i\left(x_{i} \mid \boldsymbol{f};\mathbf{w}\right)$ is the estimated probability of occupancy code $x_{i}$ at node $n_i$, which is defined in Eq. (\ref{qx}).
\section{Experiments}
\subsection{Datasets}
\subsubsection{LiDAR Dataset}
 SemanticKITTI \cite{semantickiiti} is a large sparse LiDAR dataset for self-driving. It is collected from a Velodyne HDL-64E sensor and contains 43552 scans with 4549 million points. Following VoxelContext-Net \cite{que2021VoxelContext}, we normalize the raw data into $[-1,1]^{3}$ as reference point cloud and use sequences 00 to 10 (including 23201 scans) for training, and sequences 11 to 21 (including 20351 scans) for testing.

\subsubsection{Object Point Cloud Dataset}
Microsoft Voxelized Upper Bodies (MVUB) \cite{MVUB} is a dynamic voxelized point cloud dataset containing five half-body subjects sequences with 9 and 10-bit precision. 8i Voxelized Full Bodies (MPEG 8i) \cite{MPEG8i} includes sequences of smooth surface and complete human shape point clouds with 10 and 12-bit precision. Following the setting of VoxelDNN \cite{voxeldnn}, we use Andrew10, David10, Sarah10 sequences in MVUB, Soldier10 and Longdress10 sequences in MPEG 8i for training. We select several point cloud sequences with different resolutions for testing. All testing point clouds were not used during training.
\begin{figure*}
\centering
 \subfigure {\includegraphics[scale=0.345]{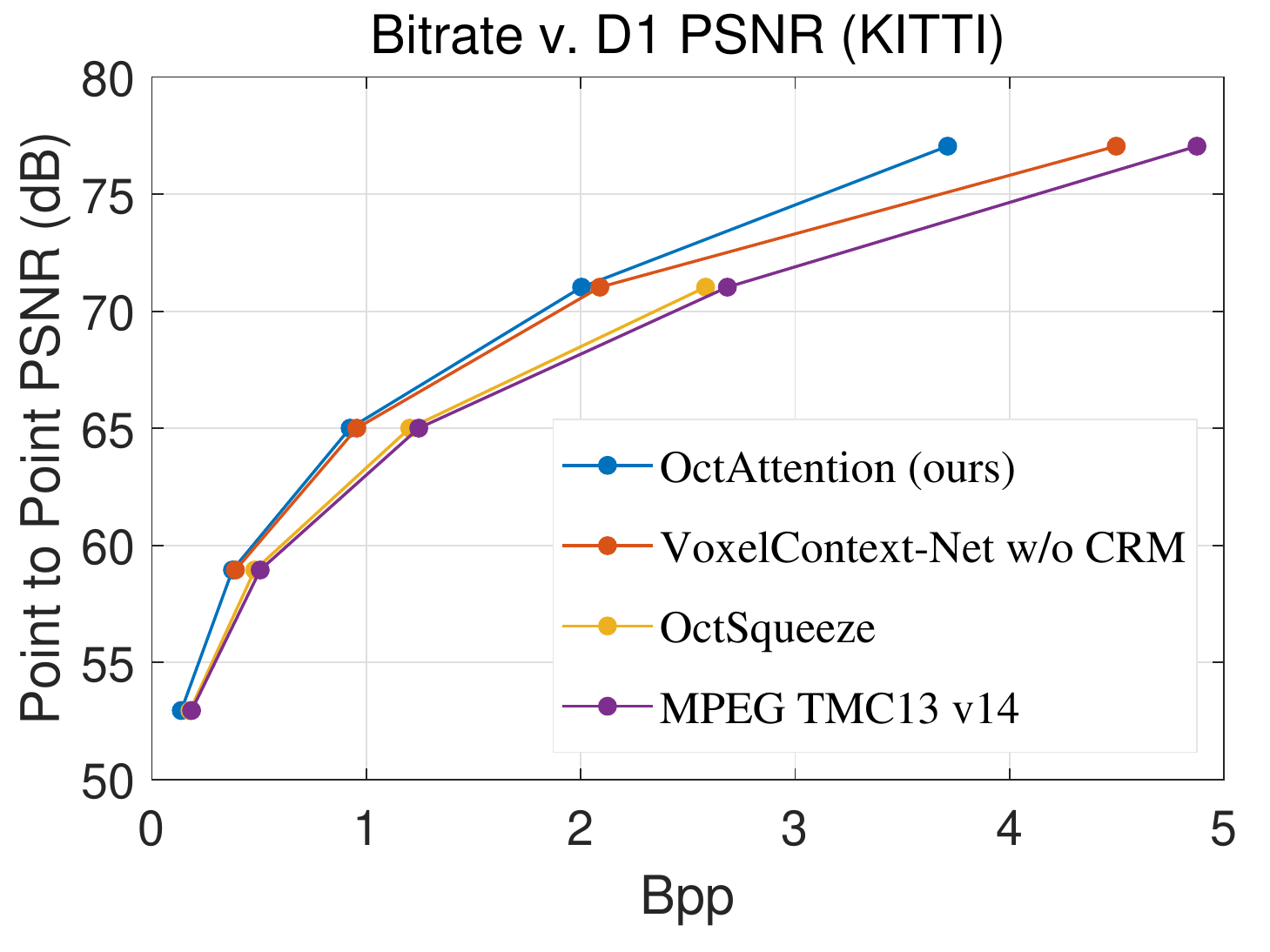}}
 \subfigure {\includegraphics[scale=0.345]{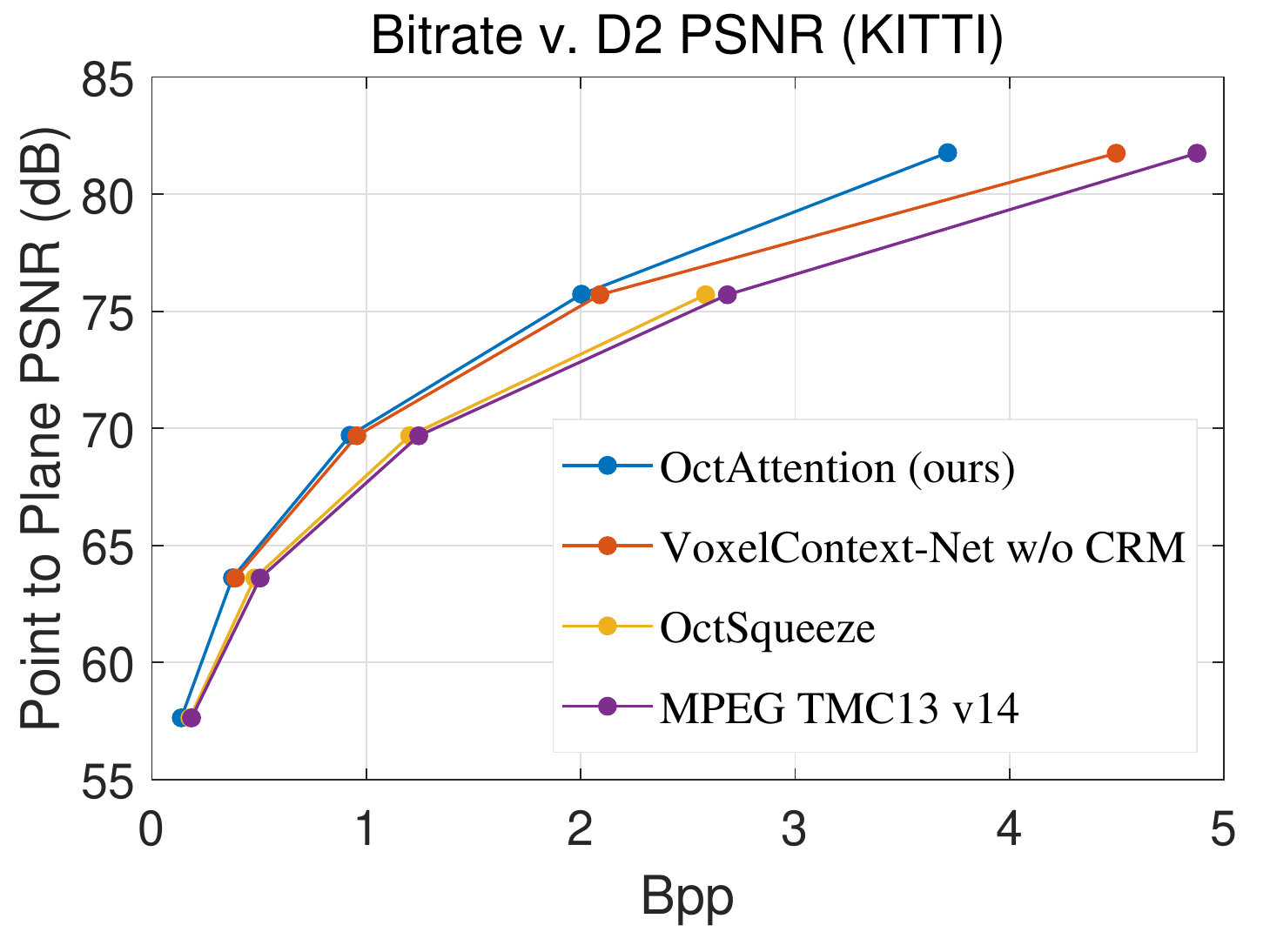}}
 \subfigure {\includegraphics[scale=0.345]{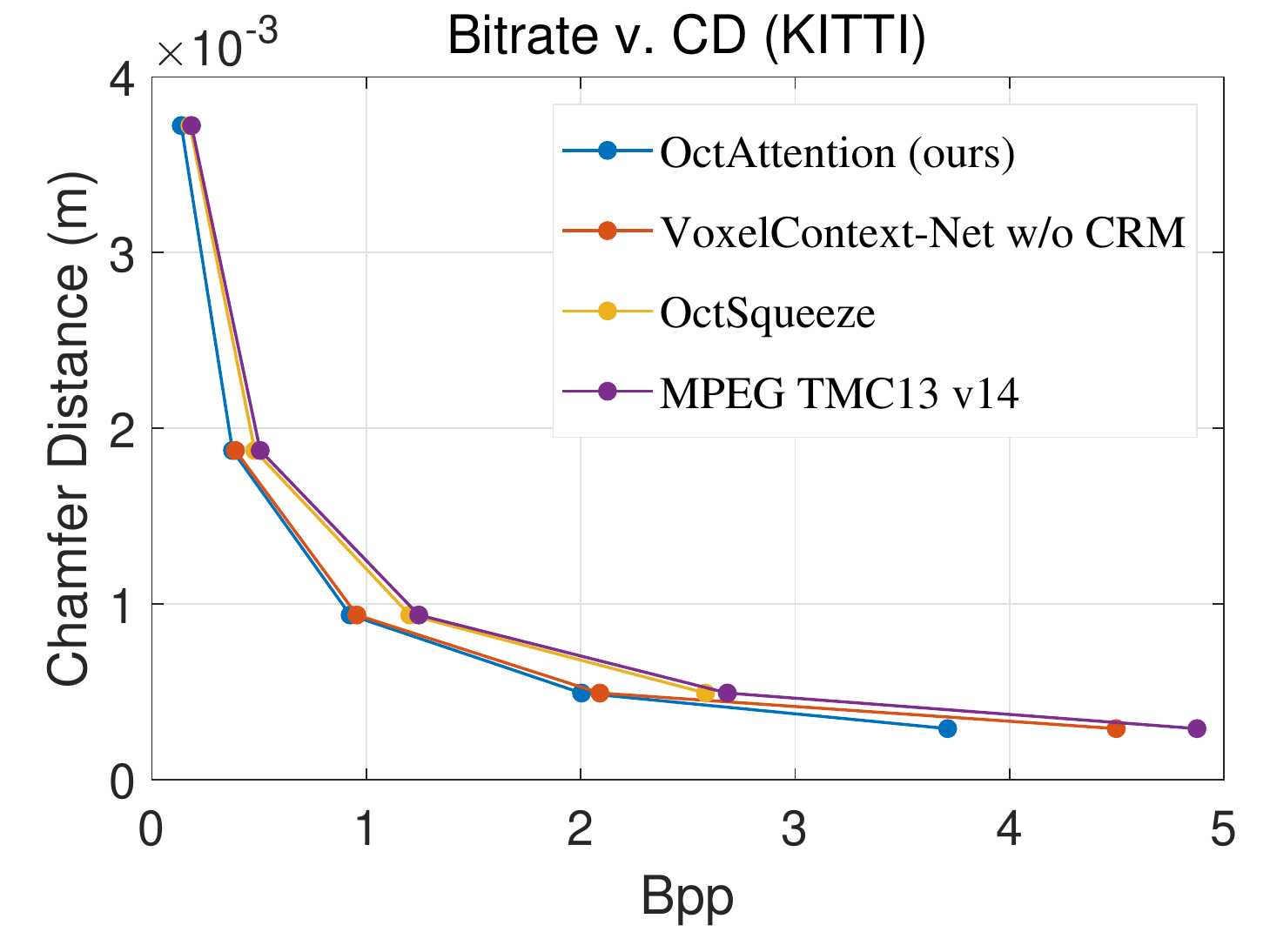}}
 \caption{Experimental results of different methods on SemanticKITTI at different bitrates.}
\label{FIGKITTI}
\end{figure*}

\begin{table*}[ht]
\renewcommand\arraystretch{0.97}
\begin{tabular}{|c|c|c|c|c|c|c|c|}
\hline
Point   Cloud  & Frames & P(full) & G-PCC     & VoxelDNN & MSVoxelDNN & \begin{tabular}[c]{@{}c@{}}OctAttention\\      (ours)\end{tabular} & \begin{tabular}[c]{@{}c@{}}Gain over \\      G-PCC\end{tabular} \\ \hline
\multicolumn{8}{|c|}{Microsoft Voxelized Upper Bodies (MVUB)}                                                                                                                                                          \\ \hline
Phil9          & 245              & 1.43    & 1.23      & 0.92     & -          & \textbf{0.83}                                                      & -32.60\%                                                        \\ \hline
Phil10         & 245              & -       & 1.07      & 0.83     & 1.02       & \textbf{0.79}                                                      & -25.97\%                                                        \\ \hline
Ricardo9       & 216              & 1.34    & 1.04      & 0.72     & -          & \textbf{0.72}                                                      & -31.20\%                                                        \\ \hline
Ricardo10      & 216              & -       & 1.07      & 0.75     & 0.95       & \textbf{0.72}                                                      & -32.53\%                                                        \\ \hline
Average        & -                & 1.34    & 0.95      & 0.81     & 0.99       & \textbf{0.76}                                                      & -30.58\%                                                        \\ \hline
\multicolumn{8}{|c|}{8i Voxelized Full Bodies (MPEG 8i)}                                                                                                                                                               \\ \hline
Loot10         & 300              & 1.03    & 0.95      & 0.64     & 0.73       & \textbf{0.62}                                                      & -35.12\%                                                        \\ \hline
Redandblack10  & 300              & 1.23    & 1.09      & 0.73     & 0.87       & \textbf{0.73}                                                      & -32.84\%                                                        \\ \hline
Boxer9/10      & 1                & -       & 0.96/0.94 & 0.76/-   & -/0.70     & \textbf{0.60/0.59}                                                 & -37.94\%                                                        \\ \hline
Thaidancer9/10 & 1                & -       & 0.99/0.99 & 0.81/-   & -/0.85     & \textbf{0.64/0.65}                                                 & -34.51\%                                                        \\ \hline
Average        & -                & 1.13    & 0.99      & 0.73     & 0.79       & \textbf{0.64}                                                      & -35.10\%                                                        \\ \hline
\end{tabular}
\centering
\caption{Average bits per point (bpp) results of different methods on MVUB and MPEG 8i.}
\label{FIGObj}

\end{table*}

\subsection{Experimental Details}
\subsubsection{Baseline}

In the static LiDAR lossy compression experiment, we quantize the point cloud $P$ by Eq. (\ref{PQ}), and set $\mathrm{L}$ from 8 to 12 to perform RD control. We compare our method against state-of-the-art methods VoxelContext-Net \cite{que2021VoxelContext} and Octsqueeze \cite{huang2020octsqueeze} in LiDAR compression. Since the source codes of the above methods are not publicly available, we keep our training/testing setting consistent with VoxelContext-Net and use the results in the paper.

In object point cloud compression, we set $\mathrm{qs}=1$ in Eq. (\ref{PQ}) to perform lossless compression. We compare our method with the hand-crafted inter-frame octree-based contexts model P(full) \cite{P(full)}, state-of-the-art compression method VoxelDNN \cite{voxeldnn} and its fast version MSVoxelDNN \cite{32}. We set the training condition following VoxelDNN and test the models on different depth data. In addition, we also compare against the most usual hand-crafted methods:  G-PCC from MPEG standard in the latest version (TMC13 v14.0) \cite{mpeg2021tmc13}.
\subsubsection{Training and Testing Strategy}
For the static LiDAR compression, we train a single model with the max octree depth of 12 as our model can learn the distribution of all layers in one model. While testing, we truncate the octree over 8-12 levels to evaluate our model at different bitrates. For object dataset compression, we train one model using the octree sequence data converted from point clouds at depths 9 and 10. We also evaluate our model on data with different geometry precision to verify robustness. We implement our model in PyTorch and perform the training/testing with Xeon E5-2637 CPU and one NVIDIA TITAN Xp GPU (12G memory). We use batch sizes of 32, epochs of 8 and Adam optimizer with a learning rate of 1e$-3$. It takes 2 days to train our model in each experiment. Occupancy, level index, and octant index are embedded into 128, 6, and 4 dimensions, respectively. We set $K=4$, $N=N_0=1024$ and use 2 layers and 4 heads in multi-head self-attention in experiments unless otherwise specified.

\subsubsection{Evaluation Metrics}
It is important to adopt the same evaluation metrics to make a fair comparison. Following the MPEG standards \cite{23}, we use two standard metrics which measure geometry reconstruction quality named point to point PSNR (D1 PSNR) and point to plane PSNR (D2 PSNR) in lossy geometry compression. Both of them can be calculated by the MPEG tool \emph{pc\_error}. We estimate the normal at each point using the MATLAB function \verb+pcnormals+. We also report chamfer distance (CD) and set PSNR peak value $r=1$ following VoxelContext-Net. We correct its results from correspondence with the authors by eliminating inconsistencies in the PSNR formula. As for the object dataset, we adopt the official default configuration in TMC13. We use bits per point (bpp) to measure the performance. Unless otherwise specified, all distortion curves and bitrates are obtained by averaging over sequence.

\subsection{Experiment Results}
\subsubsection{Results for Static LiDAR Compression}
The rate-distortion curves of LiDAR compression are shown in Fig. \ref{FIGKITTI}. We compare our method with VoxelContext-Net without coordinate refinement model (\emph{i.e.}, VoxelContext-Net w/o CRM) for fairness, as post-processing is irrelevant to compression performance evaluation. Our method outperforms other baselines at all bitrates. On average, our approach (\emph{i.e.}, OctAttention) saves 25.4\% bitrates on  SemanticKITTI compared with G-PCC, while OctSqueeze only saves less than 4\% bitrates. Our method achieves more than 11\% relative reduction in bitrate versus the state-of-the-art method VoxelContext-Net at high bitrates. It may be due to the voxel-based method failing in the sparse scenario that lacks occupied voxels. The experiment results demonstrate the effectiveness of our large receptive field context model.

\subsubsection{Results for Object Point Cloud Compression}
In Table \ref{FIGObj}, we provide the bpp results for lossless compression on object point clouds. Our method outperforms VoxelDNN and achieves a 32.8\% gain over G-PCC on average.
\begin{figure*}[ht]
\centering
\subfigure {\includegraphics[scale=0.26]{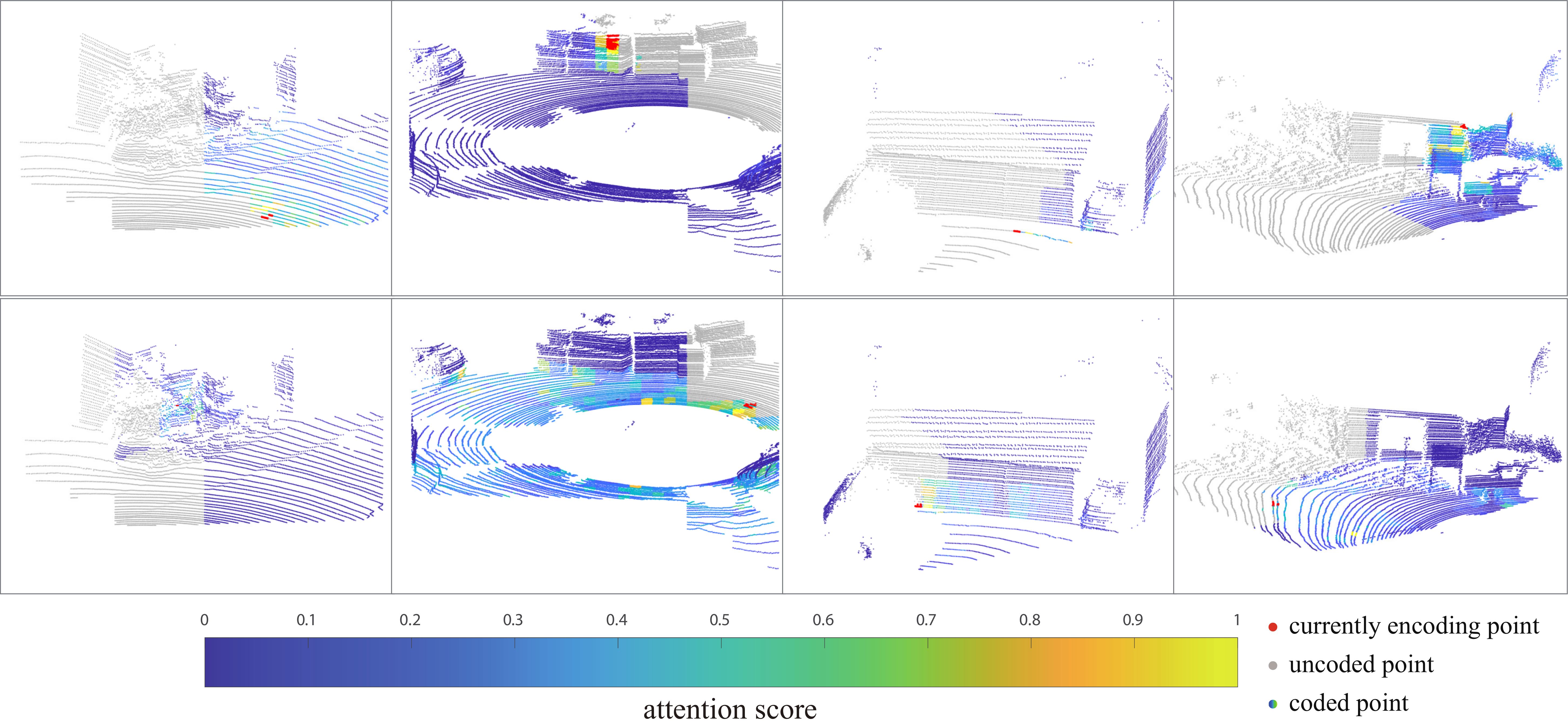}}
 \caption{Attention score visualization. Each subfigure is a context window with size $N= 1024$. Points in the currently encoding octree node ${i}$ voxel are colored in red, and other points in node $k$ are colored as their normalized attention score $a_{i,k}$ to node ${i}$. Unavailable points (\emph{i.e.}, they have not been coded yet) are colored in grey.} \label{AttMap}
\end{figure*}

\subsection{Ablation Study and Analysis}

\subsubsection{Context Window Length}
We perform an ablation experiment on SemanticKITTI to demonstrate the effectiveness of a large receptive field context. We set $N_0=N$ so that each node is predicted only once and the average receptive field is $(N+1)/2$. We then alter the context window size $N$, and the number of parameters in our model remains unchanged. As shown in Table \ref{Ablation1},  we can save 14\% bitrates by enlarging the context window size from 8 to 1024. The encoding time decreases with increased context window size due to our \emph{mask} operation, where we decrease I/O by $N_0$ times.  Decoding time does not increase significantly since the time consumption is primarily in I/O. To balance the decoding time and performance, we set $N=1024$.

\begin{table}
\begin{tabular}{|c|c|c|c|c|c|}
\hline
\multirow{2}{*}{Size $N$} & \multicolumn{3}{c|}{Bpp on SemanticKITTI}                         & \multicolumn{2}{c|}{time / K nodes (s)} \\ \cline{2-6} 
                          & D=8            & D=10           & D=12           & encode                  & decode        \\ \hline
8                         & 0.160          & 1.086          & 4.371          & 0.1628                  & 1.5091        \\ \hline
16                        & 0.158          & 1.081          & 4.334          & 0.0828                  & 1.5218        \\ \hline
32                        & 0.148          & 1.008          & 4.069          & 0.0387                 & 1.5272        \\ \hline
256                       & 0.142          & 0.959          & 3.823          & 0.0055                  & 1.5336        \\ \hline
512                       & 0.142          & 0.956          & 3.806          & 0.0028                  & 1.5949        \\ \hline
1024                      & \textbf{0.139} & \textbf{0.939} & \textbf{3.740} & \textbf{0.0015}         & 1.6093        \\ \hline
\end{tabular}
\centering
\caption{Performance and runtime when using various context window size $N$. `D' stands for the max octree depth and `time' is the duration for coding per 1000 octree nodes.}
\label{Ablation1}
\end{table}

\begin{table}
 \setlength{\tabcolsep}{1.2mm}{
\begin{tabular}{|c|c|c|c|c|c|c|}
\hline
\multirow{2}{*}{Backbone}  & \multirow{2}{*}{Sibling} & \multicolumn{5}{c|}{Bpp on SemanticKITTI}          \\ \cline{3-7} 
                           &                & D=8        & D=9          & D=10          & D=11          & D=12          \\ \hline
\multirow{2}{*}{MLP}       &                & 0.195      & 0.533        & 1.315         & 2.847         & 5.246         \\ \cline{2-7} 
                           & $\checkmark$   & 0.163      & 0.450        & 1.117         & 2.438         & 4.512         \\ \hline
\multirow{2}{*}{Attention} &                &0.197 	     &0.530 	    &1.289 	        &2.767 	        &5.073          \\ \cline{2-7} 
                           &  $\checkmark$  & \textbf{0.142}& \textbf{0.391}& \textbf{0.959}& \textbf{2.074}& \textbf{3.823}        \\ \hline
\end{tabular}}
\centering
\caption{Performance ablation study on backbone and sibling context. Backbone `MLP' stands for the shared multi-layer perception architecture like OctSqueeze.}
\label{Ablation2}
\end{table}

\subsubsection{Effectiveness of Attention and Sibling Context}
As the visualization in Fig. \ref{AttMap}, the attention mechanism discovers the similarity among points in a context window according to geometry patterns such as line, plane, surface, and curvature. The node with the highest attention score to the currently encoding nodes (red points) is colored in yellow. It confirms that the attention mechanism can predict occupancy by integrating similar features from sibling nodes in a large-scale context. In Table \ref{Ablation2}, we set $N=256$ and removed the attention and sibling features respectively for the ablation study. The results further illustrate the effectiveness of attention mechanism and sibling-involved context.

\begin{figure}
\centering
\subfigure {\includegraphics[scale=0.28]{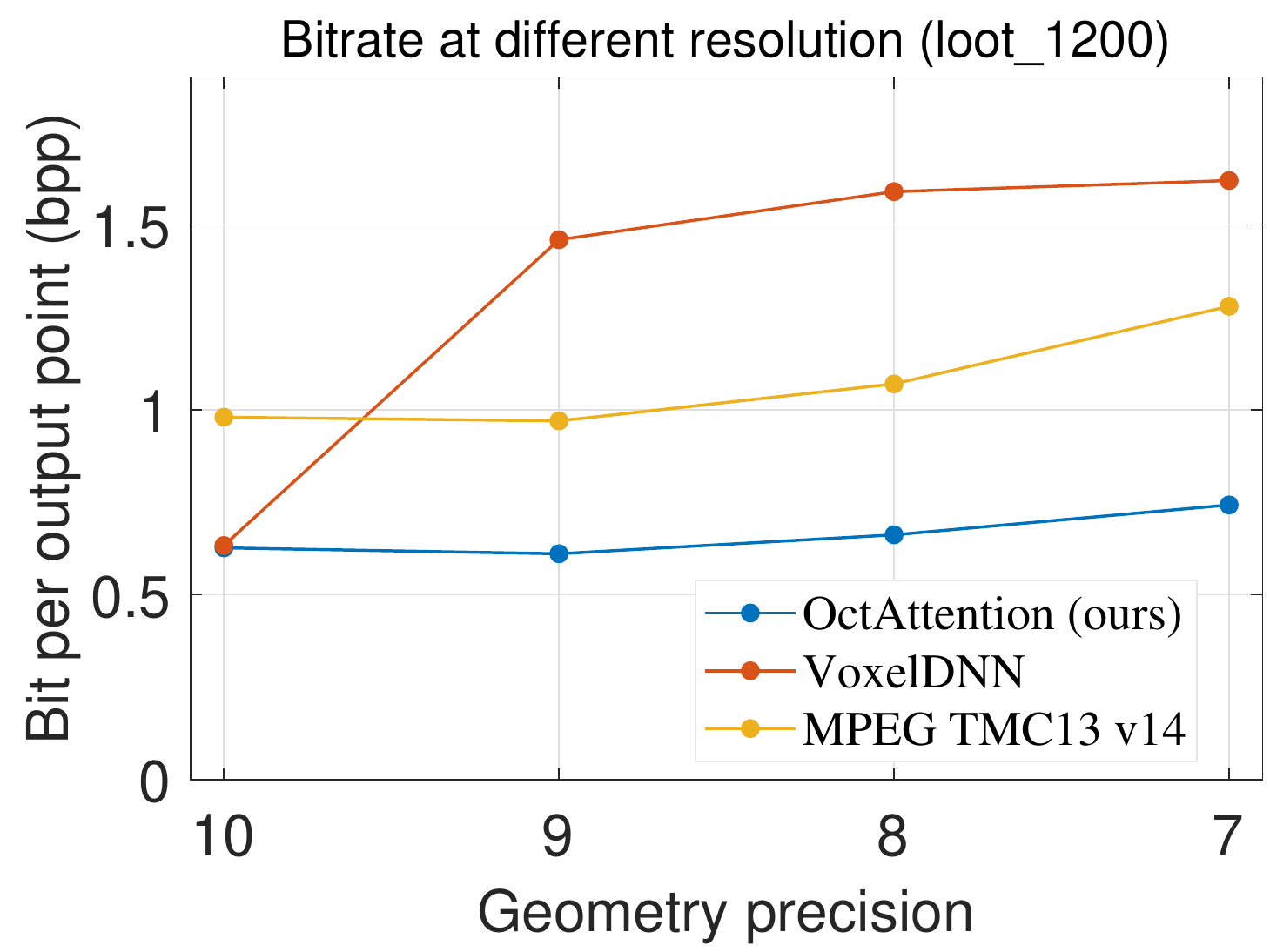}}
 \subfigure {\includegraphics[scale=0.28]{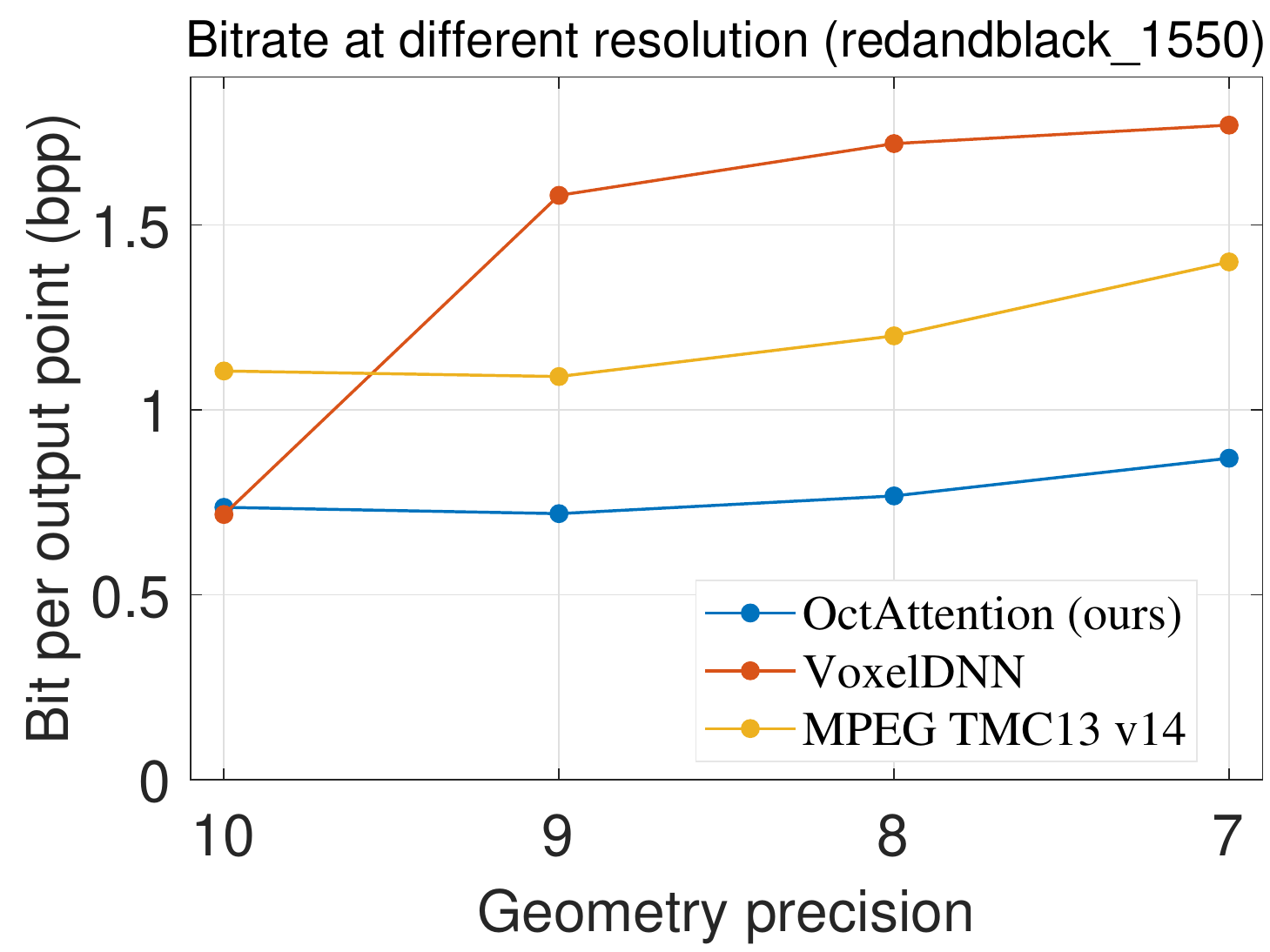}}
 \caption{Performance at different geometry precision.} \label{DenMap}
\end{figure}

\subsubsection{Robustness to Varying Point Densities}
 To evaluate \emph{OctAttention} robustness to varying point densities, we test point clouds with additional geometry precisions and point densities. See Fig. \ref{DenMap}. VoxelDNN only outperforms G-PCC on the geometry precision of 10. Its performance drops on sparse point clouds due to the lack of occupied voxels in the context. Since we adopt an octree structure with a fixed-length context window, our method's performance is shown to be stable with varying point densities.

\begin{table}

{
\setlength{\tabcolsep}{1.1mm}
\begin{tabular}{|c|c|c|c|c|}
\hline
\multicolumn{5}{|c|}{Average encoding time (s)}                                           \\ \hline
Methods & G-PCC & VoxelDNN & \begin{tabular}[c]{@{}c@{}}MSVoxelDNN\end{tabular} & Ours \\ \hline
MPEG 8i & 0.35  & 2459     & 54                                                    & 0.41 \\ \hline
MVUB    & 0.31  & 4124     & 85                                                    & 0.38 \\ \hline
\multicolumn{5}{|c|}{Average decoding time (s)}                                           \\ \hline
Methods & G-PCC & VoxelDNN & \begin{tabular}[c]{@{}c@{}}MSVoxelDNN\end{tabular} & Ours \\ \hline
MPEG 8i & 0.19  & 6274     & 58                                                    & 769  \\ \hline
MVUB    & 0.16  & 10332    & 92                                                    & 659  \\ \hline
\end{tabular}
}
\centering
\caption{Coding time (in $seconds$) on object point clouds.}
\label{coding time}
\end{table}

\subsubsection{Runtime}
The number of parameters in our model is 2.67M. See table \ref{coding time}. Our method saves 95\% encoding time and 91\% decoding time compared with VoxelDNN \cite{voxeldnn}. Our approach can be applied to real-time point cloud encoding and offline point cloud decompression. We believe it is possible to speed up the decoding by dividing the octree into disjoint subtrees and developing a GPU-based algorithm like an arithmetic encoder.

\section{Conclusion}
We proposed a novel octree-based entropy model called \emph{OctAttention} for sparse and dense point cloud geometry compression by exploiting large-scale contexts. Specifically, we extend the context and introduce sibling nodes in the octree. We employ the 
attention mechanism to emphasize the significant nodes to utilize these abundant features. We further propose a \emph{mask} operation to achieve parallel encoding under the condition of introducing siblings in the context. We evaluate our method on both the LiDAR and object point cloud datasets. The results demonstrate that the proposed method achieves state-of-the-art on both types of datasets.

\section*{Acknowledgements}
This work was supported by the National Natural Science Foundation of China (No. 62172021, 61801303, 62031013), the Shenzhen Fundamental Research Program (GXWD20201231165807007-20200806163656003), Guangdong Basic and Applied Basic Research Foundation (2019A1515012031), Shenzhen Science and Technology Plan Basic Research Project (JCYJ20190808161805519).

\bibliography{aaai22}
\clearpage
\appendix
\twocolumn[{%
    \renewcommand\twocolumn[1][]{#1}%
    \begin{center}
        \centering
        \LARGE \textbf{\appendixname}
        \vspace{30pt}
    \end{center}%
}]
\setcounter{secnumdepth}{3}
\renewcommand\thesection{\arabic{section}.}

\makeatletter 
\renewcommand{\section}{\@startsection{section}{1}{0mm}
 {-\baselineskip}{0.5\baselineskip}{\bf\leftline}}
\makeatother

\section{Compression performance at high bitrates}
\begin{figure}[!ht]
\centering
 \subfigure {\includegraphics[scale=0.28]{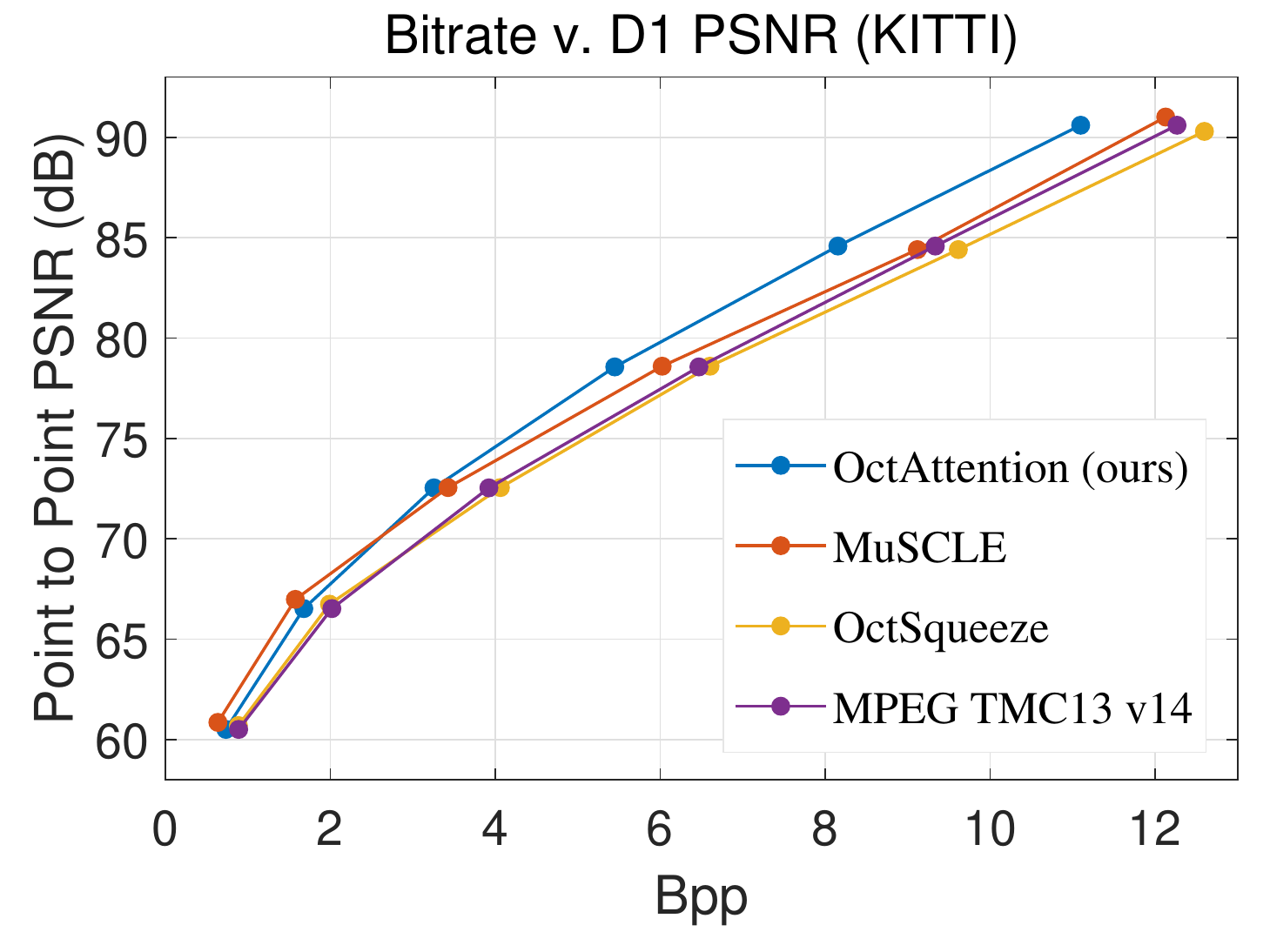}}
 \subfigure {\includegraphics[scale=0.28]{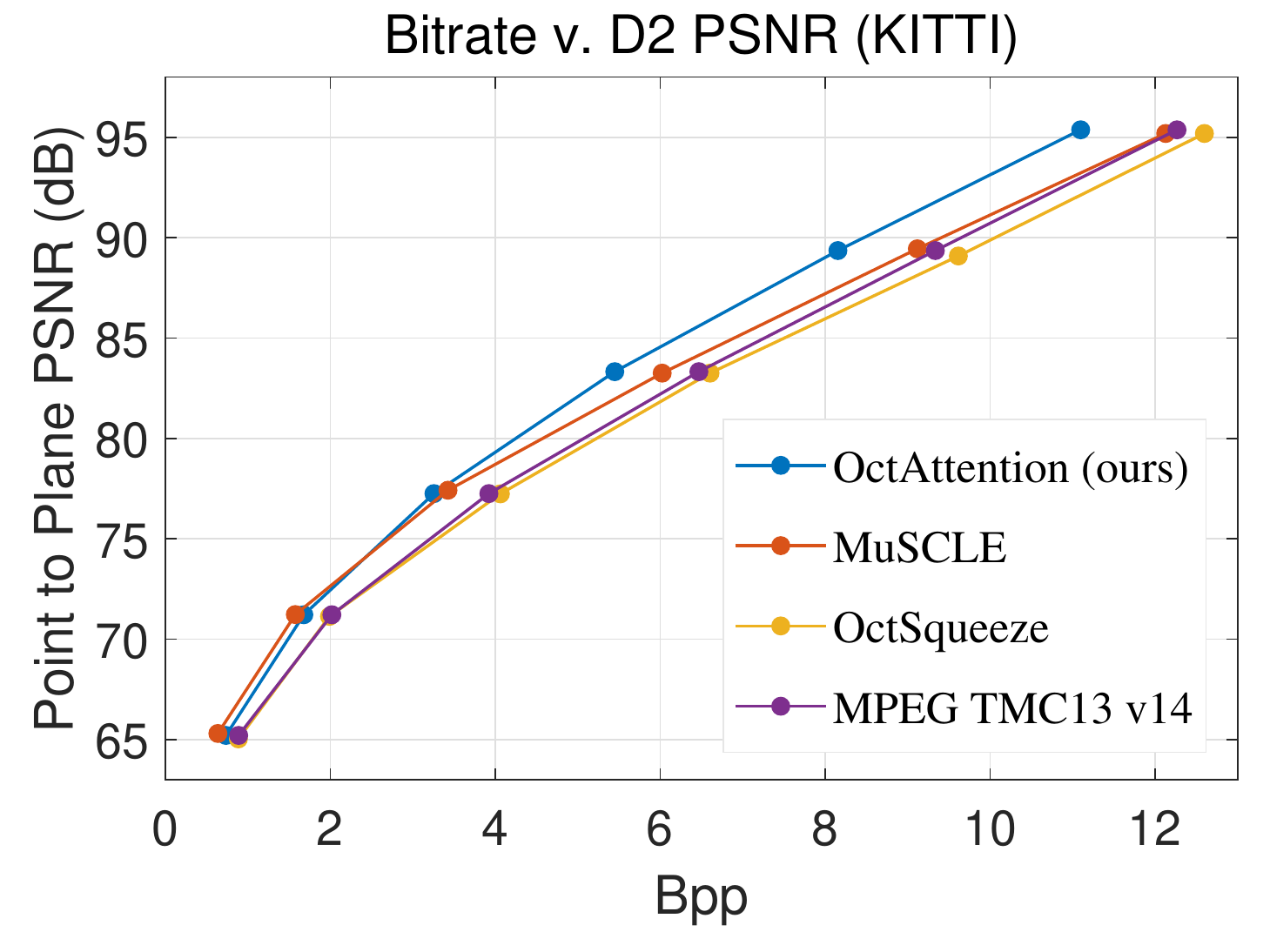}}
 \caption{Experimental results of different methods on SemanticKITTI quantized by octree with the depth of 11-16.}
\label{suppKITTI}
\end{figure}

To further validate the compression performance and robustness of our model, we quantize the point clouds with smaller $\mathrm{qs}$. Thus we obtain point clouds with minor reconstruction error and higher bitrates. Since VoxelContext-Net \cite{que2021VoxelContext} only reports its performance at low bitrates and its quantization and evaluation settings are different from OctSqueeze\cite{huang2020octsqueeze} and MuSCLE\cite{biswas2021muscle}, we follow the quantization setting in MuSCLE. We set $\text{offset}=-200$, $\mathrm{qs}=400/(2^\mathrm{L}-1)$ in Eq. (\ref{PQ}) and PSNR peak value $r=59.70$. We validate our model without retraining. The experimental results are shown in Fig. \ref{suppKITTI}.

 Our model is trained on 12 depth octrees and tested on 11-16 depth data, which is almost non-existent in the training data. As shown in Fig. \ref{suppKITTI}, our method outperforms OctSqueeze at all bitrates and saves 19.4\% bitrates on average. At the highest bitrate, our method saves 13.5\% bitrates compared with OctSqueeze and saves 8.5\% bitrates compared with MuSCLE. It should be noted that temporal correlations are utilized in MuSCLE to encode dynamic point clouds, even though OctAttention does not make use of inter-frame information. Nevertheless, we achieve comparable compression performance at low bitrates and better performance at high bitrates. The experimental results demonstrate the effectiveness and robustness of our method at high bitrates.

\section{Visualization of embedding}
Embedding is supposed to discover the local geometry patterns which are hidden in octree representation. To demonstrate its effectiveness, we visualize the embedded feature $\boldsymbol{f}_k=[\boldsymbol{h}_k^{(0)},\boldsymbol{h}^{(1)}_{k},\boldsymbol{h}^{(2)}_{k}$$,...,\boldsymbol{h}^{(K-1)}_{k}]$ of node $n_k$, where $\boldsymbol{h}^{(0)}_{k}$ represents the feature $\left[S_kW_{1}, L_kW_{2},O_kW_3\right]$ of node $n_k$ and $\boldsymbol{h}^{(1)}_{k},\boldsymbol{h}^{(2)}_{k}$$,...,\boldsymbol{h}^{(K-1)}_{k}$ represent the features of its $K-1$ ancestors. $S_k,L_k,O_k$ are one-hot coded occupancy, level index and octant index, and $W$ is their respective embedding matrix. We employ Principal Component Analysis to reduce $\boldsymbol{f}_k$ dimension to 3 and  then color corresponding points with the 3-dimension vectors. We also visualize the original feature of node $n_k$, which are defined as the concatenation of $[S_k,L_k,O_k]$ and features of its $K-1$ ancestors. We implement our experiment on ModelNet40 \cite{modelnet40}.

\begin{figure}[!ht]
\centering
 \subfigure{\includegraphics[scale=0.22]{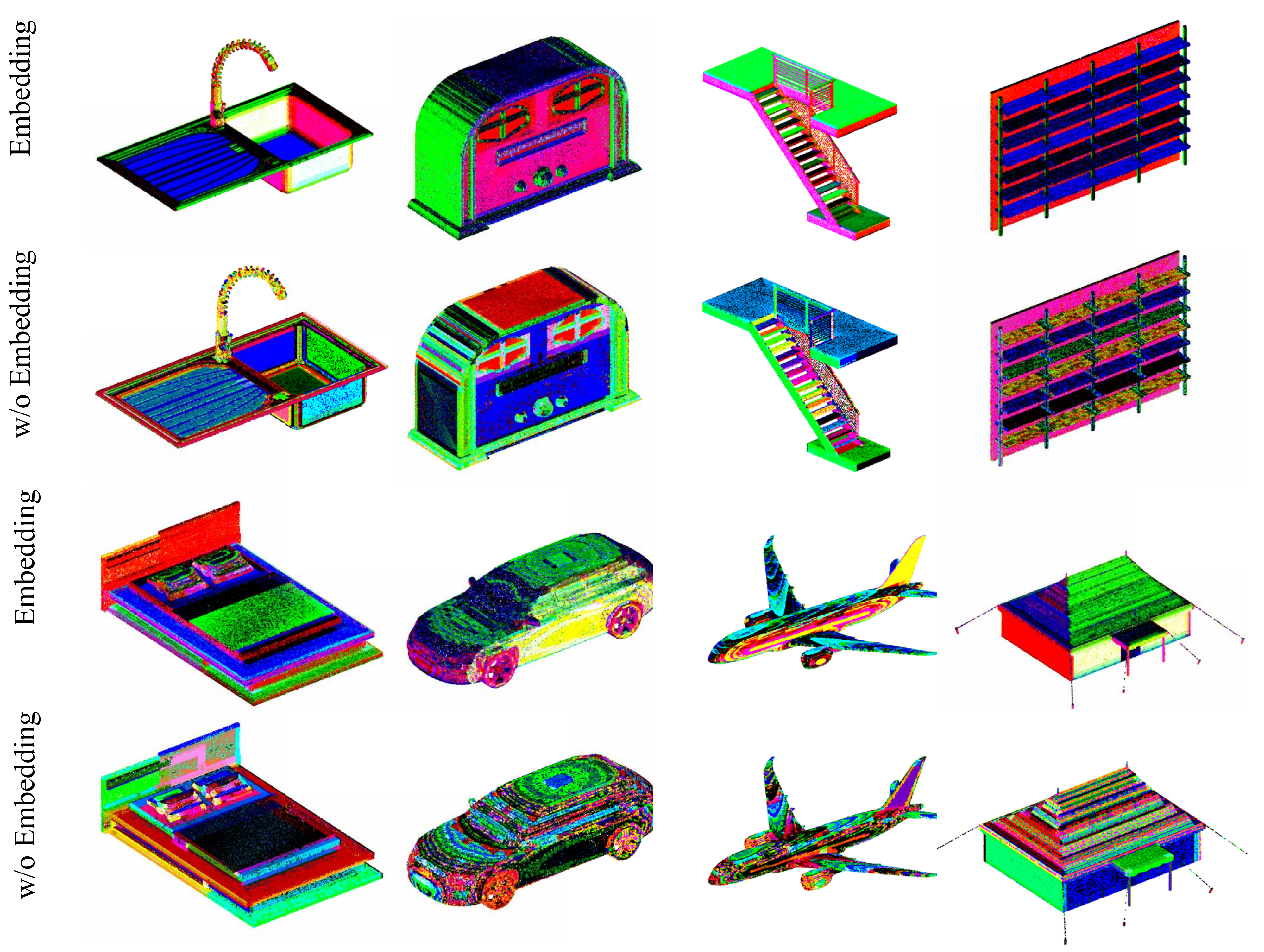}}
 \caption{Visualization of embedded features (Embedding) and original features (w/o Embedding). The color represents the feature after dimensionality reduction.}
\label{embedding}
\end{figure}

Fig. \ref{embedding} shows that embedding can describe local features such as surfaces, orientation, and curvatures. For example, original features are unable to distinguish surfaces explicitly or discover parallel planes. Points within the same surface (see tent, car, bed, radio, and airplane) or parallel planes (see sink, bookshelf, and stairs) may have varying colors, resulting from the inconsistency between the occupancy code value and geometry patterns. On the contrary, similar geometry patterns are rendered in similar colors by embedding. More specifically, the colors of the upper and lower surfaces are blue and green, and the colors of the left and right sides are yellow and red. In summary, embedding can serve as an enhanced local geometry descriptor, which is helpful to model point cloud geometry distribution.

\end{document}